\documentclass[sigconf,screen]{acmart}
\usepackage{fancyhdr}
\usepackage{flushend}
\usepackage{booktabs} % For formal tables

\setcopyright{licensedusgovmixed}
\acmPrice{15.00}
\acmDOI{10.1145/3243127.3243132}
\acmYear{2018}
\copyrightyear{2018}
\acmISBN{978-1-4503-5972-6/18/09}
\acmConference[MASES '18]{Proceedings of the 1st International Workshop on Machine Learning and Software Engineering in Symbiosis}{September 3, 2018}{Montpellier, France}
\acmBooktitle{Proceedings of the 1st International Workshop on Machine Learning and Software Engineering in Symbiosis (MASES '18), September 3, 2018, Montpellier, France}

\begin{document}
	
	\author{Ben Gelman}
	\affiliation{%
		\institution{Two Six Labs, LLC.}
		\city{Arlington} 
		\state{Virginia} 
		\country{USA}
	}
	\email{ben.gelman@twosixlabs.com}
	
	\author{Bryan Hoyle}
	\affiliation{%
		\institution{Two Six Labs, LLC.}
		\city{Arlington} 
		\state{Virginia} 
		\country{USA}
	}
	\email{bryan.hoyle@twosixlabs.com}
	
	\author{Jessica Moore}
	\affiliation{%
	\institution{Two Six Labs, LLC.}
	\city{Arlington} 
	\state{Virginia} 
	\country{USA}
	}
	\email{jessica.moore@twosixlabs.com}
	
	\author{Joshua Saxe}
	\affiliation{%
	\institution{Sophos}
	\city{Fairfax} 
	\state{Virginia}
	\country{USA}
	}
	\email{joshua.saxe@sophos.com}
	
	\author{David Slater}
	\affiliation{%
	\institution{Two Six Labs, LLC.}
	\city{Tacoma} 
	\state{Washington}
	\country{USA}
	}
	\email{david.slater@twosixlabs.com}
	
	\title{A Language-Agnostic Model for Semantic Source Code Labeling}
	

	\begin{abstract}
		Code search and comprehension have become more difficult in recent years due to the rapid expansion of available source code. Current tools lack a way to label arbitrary code at scale while maintaining up-to-date representations of new programming languages, libraries, and functionalities. Comprehensive labeling of source code enables users to search for documents of interest and obtain a high-level understanding of their contents. We use Stack Overflow code snippets and their tags to train a language-agnostic, deep convolutional neural network to automatically predict semantic labels for source code documents. On Stack Overflow code snippets, we demonstrate a mean area under ROC of 0.957 over a long-tailed list of 4,508 tags. We also manually validate the model outputs on a diverse set of unlabeled source code documents retrieved from Github, and obtain a top-1 accuracy of 86.6\%. This strongly indicates that the model successfully transfers its knowledge from Stack Overflow snippets to arbitrary source code documents.
	\end{abstract}
	
	
	

	\maketitle
	
	\section{Introduction}

In recent years, the quantity of available source code has been growing exponentially \cite{source_growth}. Code reuse at this scale is predicated on understanding and searching through a massive number of projects and source code documents. The ability to generate meaningful, semantic labels is key to comprehending and searching for relevant source code, especially as the diversity of programming languages, libraries, and code content continues to expand.

The search functionality for current large code repositories, such as GitHub \cite{github} and SourceForge \cite{sourceforge}, will match queried terms in the source code, comments, or documentation of a project. More sophisticated search approaches have shown better performance in retrieving relevant results, but they often insufficiently handle scale, breadth, or ease of use. Santanu and Prakash \cite{patterns} develop pattern languages for C and PL/AS that allow users to write generic code-like schematics. Although the schematics locate specific source code constructs, they do not capture the general functionality of a program and scale poorly to large code corpora. Bajracharya et al. \cite{sourcerer} develop a search engine called Sourcerer that enhances keyword search by extracting features from its code corpus. Sourcerer scales well to large corpora, but it is still hindered by custom language-specific parsers. Suffering from a similar problem, Exemplar \cite{exemplar} is a system that tracks the flow of data through various API calls in a project. Exemplar also uses the documentation for projects/API calls in order to match a user's keywords. Recent applied works have similar shortcomings \cite{openhub} \cite{krugle} \cite{searchcode} \cite{sourcegraph}. Creating a solution that operates across programming languages, libraries, and projects is difficult due to the complexity of modeling such a huge variety of code.

\begin{figure}[t]
	\centering
	\includegraphics[width=\linewidth]{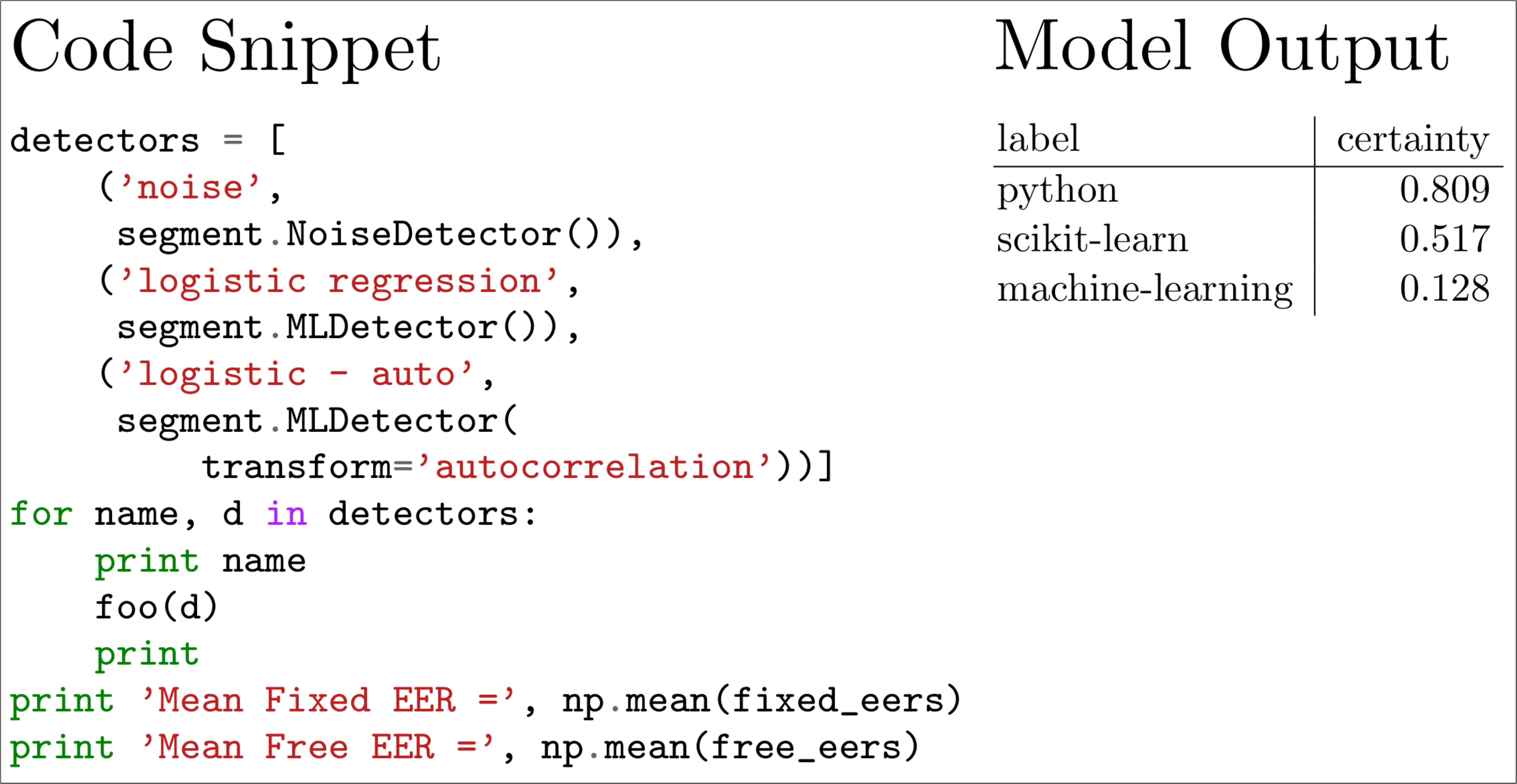}
	\caption{An example prediction of our model. The input code snippet is on the left, while the predicted labels and their raw certainties are on the right. Keyword matching on the predicted labels would not have been able to locate this code. }
	\label{fig:code_snippet}	
\end{figure}

As a step in that direction, we present a novel framework for generating labels for source code of arbitrary language, length, and domain. Using a machine learning approach, we capitalize on a wealth of crowdsourced data from Stack Overflow (SO) \cite{stackoverflow}, a forum that provides a constantly growing source of code snippets that are user-labeled with programming languages, tool sets, and functionalities. Prior works have attempted to predict a single label for an SO post \cite{coocc_pred} \cite{predicting_tags} using both the post's text and source code as input. To our knowledge, our work is the first to use Stack Overflow to predict exclusively on source code. Additionally, prior methods do not attempt multilabel classification, which becomes a significant issue when labeling realistic source code documents instead of brief SO snippets. Our approach utilizes SO's code snippets to simultaneously model thousands of concepts and predict on previously unseen source code, as demonstrated in Fig. \ref{fig:code_snippet}.

We construct a deep convolutional neural network that directly processes source code documents of arbitrary length and predicts their functionality using pre-existing Stack Overflow tags. As users ask questions about new programming languages and tools, the model can be retrained to maintain up-to-date representations. 

Our contributions are as follows:
\begin{itemize}
	\item First work, to our knowledge, to introduce a baseline for multilabel tag prediction on Stack Overflow posts. 
	\item Convolutional neural network architecture that can handle arbitrary length source code documents and is agnostic to programming language.
	\item State-of-the-art top-1 accuracy (79\% vs 65\% \cite{predicting_tags}) for predicting tags on Stack Overflow posts, using only code snippets as input.
	\item Approach that enables tagging of source code corpora external to Stack Overflow, which is validated by a human study.
\end{itemize}

We organize the rest of the paper as follows:  section \ref{related} discusses related works, section \ref{data} details data preprocessing and correction, section \ref{methodology} explains our neural network architecture and validation, section \ref{results} displays our results, section \ref{challenges} presents challenges and limitations, and section \ref{conclusions} considers future work. 

\section{Related Work} \label{related}

Due to the parallels between source code and natural language \cite{naturalness} \cite{surveybigcodenatural}, we find that recent work in the natural language processing (NLP) domain is relevant to our problem. Modern NLP approaches have generated state-of-the-art results with long short-term memory neural networks (LSTMs) and convolutional neural networks (CNNs). Sundermeyer, Schl{\"u}ter, and Ney \cite{lstm_language} have shown that LSTMs perform better than n-grams for modeling word sequences, but the vocabulary size for word-level models is often large, requiring a massive parameter space. Kim, Jernite, Sontag, and Rush \cite{character_aware} show that by combining a character-level CNN with an LSTM, they can achieve comparable results while having 60\% fewer parameters. Further work shows that CNNs are able to achieve state-of-the-art performance without the training time and data required for LSTMs \cite{nn_models}. In the source code domain, however, prior work has utilized a wide variety of methods.

In 1991, Maarek, Berry, and Kaiser \cite{ir_libraries} recognized that there was a lack of usable code libraries. Libraries were difficult to find, adapt, and integrate without proper labeling, and locating components functionally close to a given topic posed a challenge. The authors developed an information retrieval approach leveraging the co-occurrence of neighboring terms in code, comments, and documentation. 

More recently, Kuhn, Ducasse, and G{\'\i}rba \cite{semantic_clustering} apply Latent Semantic Indexing (LSI) and hierarchical clustering in order to analyze source code vocabulary without the use of external documentation. LSI-based methods have had success in the code comprehension domain, including document search engines \cite{lsi_search} and IDE-integrated topic modeling \cite{relational_topics}. Although the method seems to perform well, labeling an unseen source code document requires reclustering the entire dataset. This is a significant setback for maintaining a constantly growing corpus of labeled documents. 

In the context of source code labeling, supervised methods are mostly unexplored. A critical issue in this task is the massive amount of labeled data required to create the model. A few efforts have recognized Stack Overflow for its wealth of crowdsourced data. Saxe, Turner, and Blokhin \cite{crowd} search for Stack Overflow posts containing strings found in malware binaries, and use the retrieved tags to label the binaries. Kuo \cite{coocc_pred} attempts to predict tags on SO posts by computing the co-occurrence of tags and words in each post. He achieves a 47\% top-1 accuracy, which in this context is the task of predicting only one tag per post.

Clayton and Byrne \cite{predicting_tags} also attempt to predict a tag for SO posts. They invest a great deal of effort in feature extraction inspired by ACT-R's declarative memory retrieval mechanisms \cite{act-r}. Utilizing logistic regression, they achieve a 65\% top-1 accuracy.

In this work, we generate a more complex machine learning model than those present in previous attempts. Because we intend to generalize our model to source code files, we make our tests stricter by only using the actual code inside Stack Overflow posts as inputs to the model. Despite the information loss from not taking advantage of the entire post text, we still further improve on the performance of prior work and obtain a 78.7\% top-1 accuracy.

\section{Data} \label{data}

The primary goal of our work is to create a machine learning system that will classify the functionality of source code. We achieve this by leveraging Stack Overflow, a large, public question and answer forum focused on computer programming. Users can ask a question, provide a response, post code snippets, and vote on posts. The SO dataset provides several advantages in particular: a huge quantity of code snippets; a wide set of tags that cover concepts, tools, and functionalities; and a platform that is constantly updated by users asking and answering questions about new technologies. Due to the complexity of the data, we use this section to discuss the data's characteristics and our preprocessing procedures in detail. 

\begin{figure}[t]
	\centering
	\includegraphics[width=.95\linewidth]{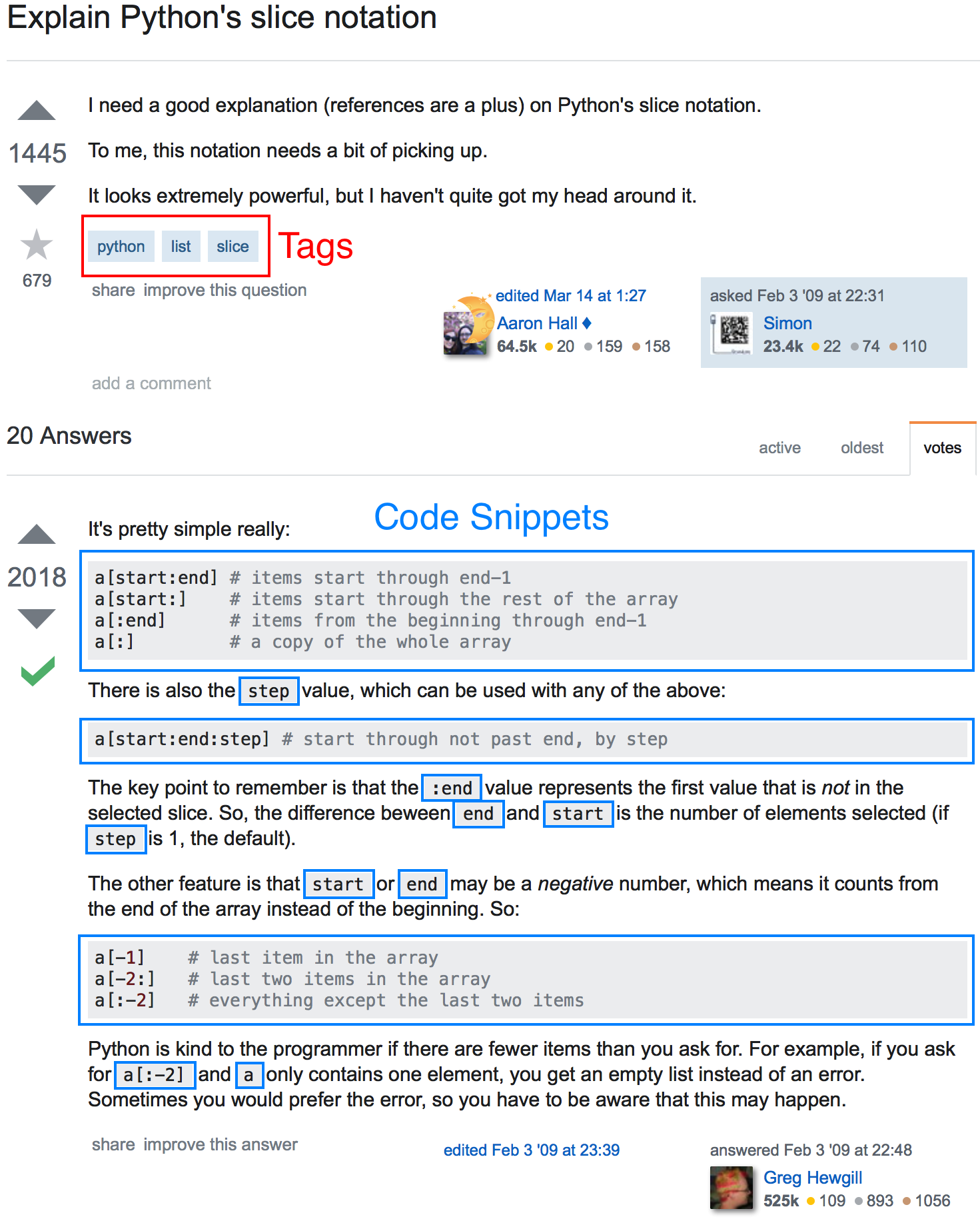}
	\caption{A Stack Overflow thread with a question and answer. The thread's tags are boxed in red and the code snippets are boxed in blue. For the purpose of training our model, the tags are the output labels and the code snippets are the input features. We can see from this example that the longer snippets look like valid code, while the shorter snippets are not as useful.}
	\label{fig:so_post}
\end{figure}

Fig. \ref{fig:so_post} is an example of a Stack Overflow thread. Users who ask a question are allowed to attach a maximum of five tags. Although there is a long list of previously used tags, users are free to enter any text. The tags are often chosen with programming language, concepts, and functionality in mind. The tags for this example, boxed in red, are ``python," ``list," and ``slice." Additionally, any user is allowed to block off text as a code snippet. In this example, the user providing an answer uses many code snippets, which have been boxed in blue. Although the code snippets may describe a particular functionality, they do not necessarily represent a complete or syntactically correct program.

Our initial intuition is that the code snippets can simply be input into a machine learning model with the user-created tags as labels. This trained model would then be able to accept any source code and provide tags as output. As we further analyze the data, several questions need to first be resolved, including how to associate tags with snippets, what constitutes a single code sample, and which data should be filtered from the dataset.

Stack Overflow's threads are the fundamental pieces of our training data. The publicly available SO data dump provides over 70 million source code snippets with labels that would be useful for real world projects. Because the tags are selected at the thread level while snippets occur in individual posts, we assign the thread's tags to each post in that thread. Since a single post can have many code snippets, we choose to concatenate the snippets using newline characters as separators in order to preserve a user's post as a single idea.  

Although these transformations ensure that a post will suffice as input to a language-level model, they do not guarantee the usefulness of the snippets themselves. The following section will address several problems with short, uninformative code snippets, user error in tagging posts and generating code with the correct functionality, and the long-tailed distribution of unusable tags. 

\subsection{Statistics and Data Correction}

As of December 15, 2016, the Stack Overflow data dump contains 24,158,127 posts that have at least one code snippet, 73,934,777 individual code snippets, and 46,663 total tags. Despite the large amount of data, there is a severe long-tailed phenomenon that is common in many online communities \cite{onlineproperties}. The distributions of code-snippet length and number of tags per post are of particular importance to our problem. 

%As a crowd-sourced dataset, there are inevitably problems resulting from human processes and errors that are not optimal for training our machine learning model. We make corrections to the data for small snippets, down-voted posts, and very rare tags. 
%The significant quantity of small snippets raised suspicions about the validity of that data. 

\begin{figure}[t]
	\centering
	\includegraphics[width=\linewidth]{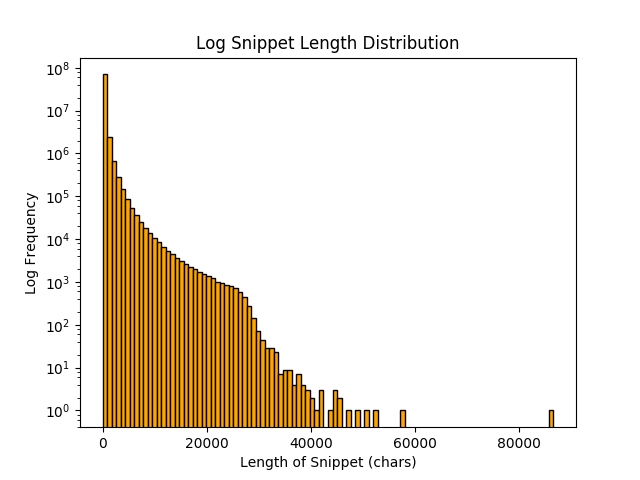}
	\caption{The distribution of snippet lengths in the full dataset, with frequencies logarithmically scaled. Although short code snippets are extremely common, they have limited value.}
	\label{fig:log_snip_length}
\end{figure}

\begin{figure}[t]
	\centering
	\includegraphics[width=\linewidth]{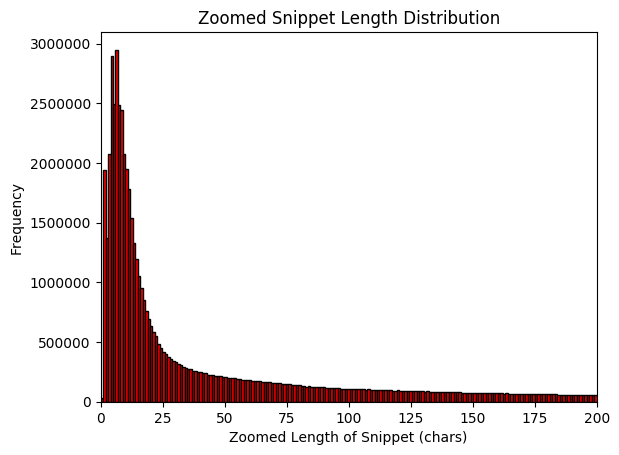}
	\caption{A zoomed view of the snippet length distribution, with 1 bin equal to 1 character. There are many strings that are empty or only a few characters long.}
	\label{fig:zoomed_snip_length}
\end{figure}

Fig. \ref{fig:log_snip_length} shows the distribution of individual snippet lengths, measured in number of characters, throughout Stack Overflow. As one would expect, the longer snippets are many orders of magnitude less frequent than the shorter snippets. Fig. \ref{fig:zoomed_snip_length} further demonstrates that, of the many short snippets, there is a huge quantity that are empty strings or are only a few characters long. There are several reasons why these snippets are poor choices for training data. First, a single character is usually not descriptive enough to characterize multiple tags. Saying that `x' is a good indicator of python, machine learning, and databases does not make sense. Going back to Fig. \ref{fig:so_post}, we can also see that the short snippets are often references to code, but not valid code themselves. 

\begin{figure}[t]
	\centering
	\includegraphics[width=\linewidth]{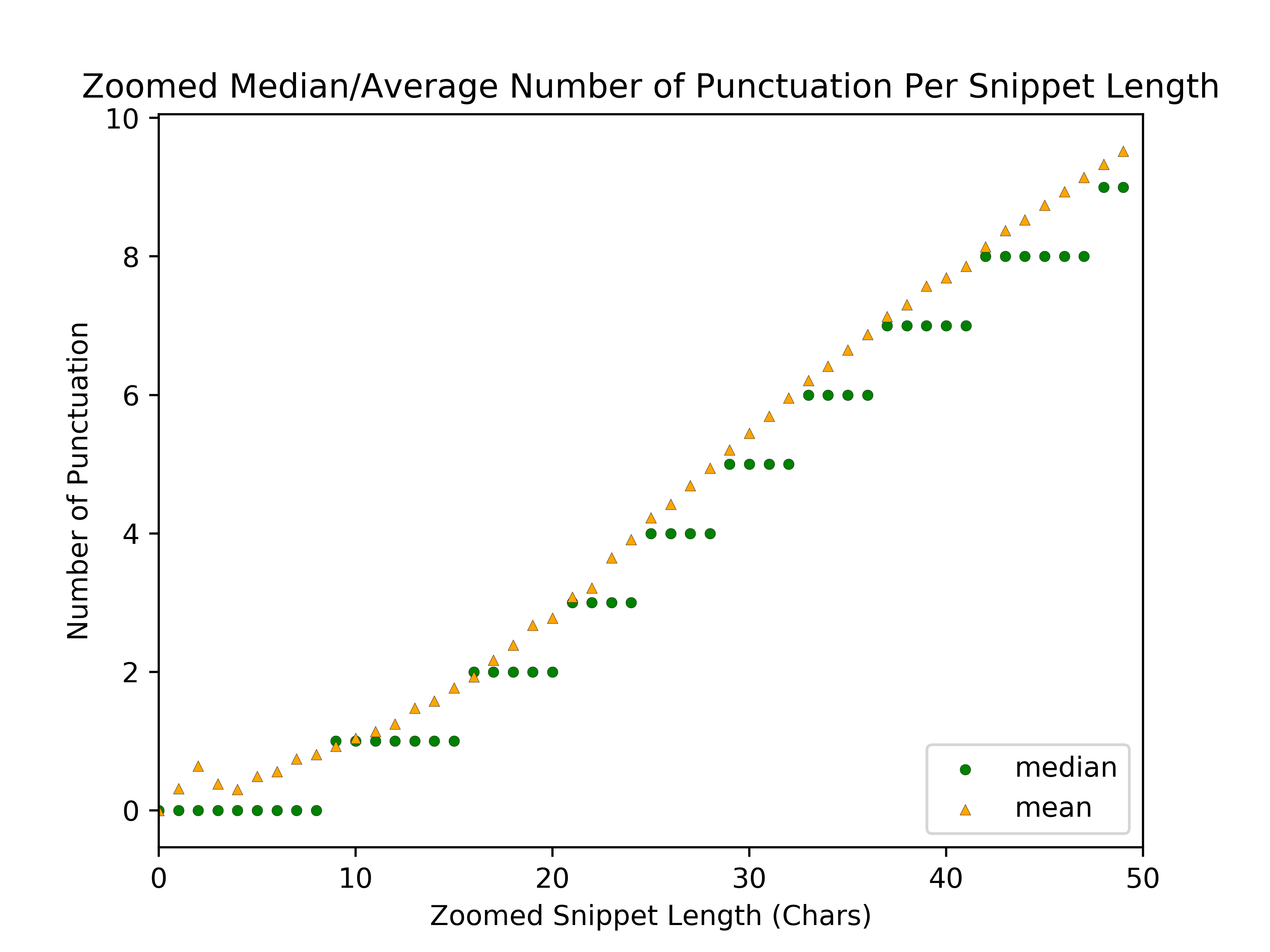}
	\caption{The mean and median number of punctuation marks at different snippet lengths. At a snippet length of 10 characters, the mean and median number of punctuation marks is 1, indicating a reasonable choice for minimum snippet length.}
	\label{fig:zoomed50_med_punctuation}
\end{figure}

In order to avoid cutting out snippets at an uninformed threshold, we investigate snippets of different lengths in more detail. We found punctuation to be a good indicator of code usefulness in short snippets. The occurrence of punctuation means that we are more likely to see programming language constructs such as ``variable = value" or ``class.method." However, simply removing all snippets without punctuation is not viable because of valuable alphanumeric keywords and punctuation-free code (``call WriteConsole''), so we instead decide to filter based on a threshold length. Fig. \ref{fig:zoomed50_med_punctuation} shows the median and mean number of punctuation marks for different snippet lengths. At a snippet length of 10 characters, the mean and median are both greater than one, so we filter out all snippets that are length 9 or below from the data. 

\begin{figure}[t]
	\centering
	\includegraphics[width=\linewidth]{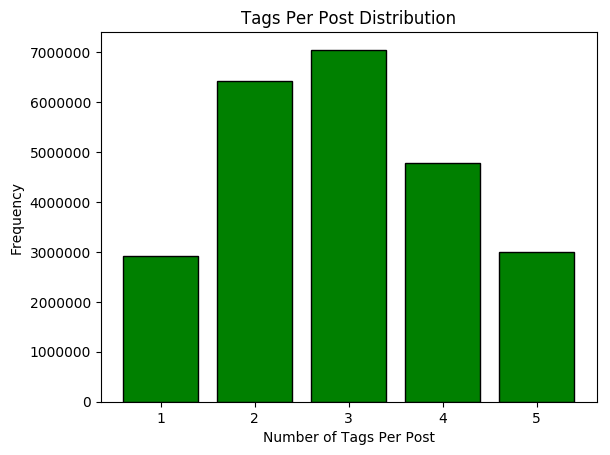}
	\caption{Distribution of tags per post. All posts on Stack Overflow must have at least one tag, but there is a maximum of five tags, resulting in missing labels. }
	\label{fig:tags_per_post}
\end{figure}

Additionally, Fig. \ref{fig:tags_per_post} shows the distribution of tags per post. As stated previously, Stack Overflow allows a maximum of five tags for any given post. Although most posts contain three tags, there is still a significant number of posts with fewer tags. The combined effect from a high quantity of posts that have few tags and an enforced maximum creates a ``missing label phenomenon." This is the situation where a given post is not tagged with \emph{all} of the functionalities or concepts actually described in the post. This is a non-trivial challenge for machine learning models because a code snippet is considered a negative example for a given label if that label is missing.

\begin{figure}[t]
	\centering
	\includegraphics[width=\linewidth]{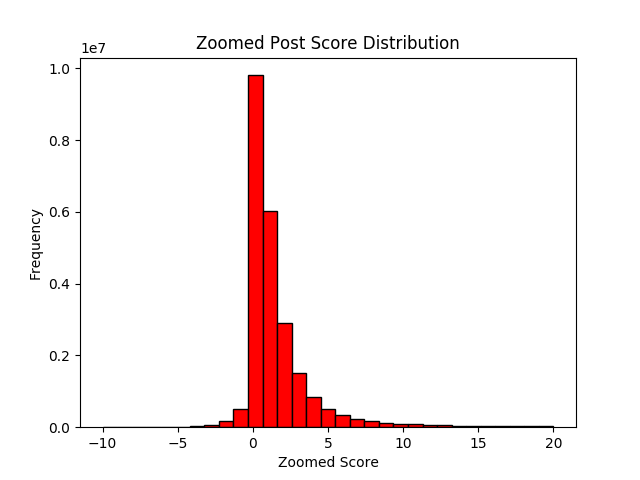}
	\caption{The distribution of scores on Stack Overflow posts. Negative scores are often the result of poorly worded questions, incorrectly tagged posts, or flawed code snippets, so we filter them out of the training set. We keep zero-scored snippets because they may not have been viewed enough to be voted on.}
	\label{fig:zoomed_scores}
\end{figure}

Users can also add errors to the training data by simply being wrong about their tags or posted code on Stack Overflow. Because users can vote based on the quality of a post, we can use scores as an indicator for incorrectly tagged or poorly coded posts. Fig. \ref{fig:zoomed_scores} shows the distribution of scores for posts that have at least one code snippet. We cut all posts with negative scores from the training data. Although we considered cutting posts with zero score because they had not been validated by other users via voting, we ultimately choose to keep them because the score distribution shows that there is a large amount of data with zero score.

\begin{table}
	\caption{Rankings are based on the number of posts that are labeled with a tag, after filtering data for snippet and score thresholds. This shows that the majority of tags have too few samples to train and validate a machine learning model. }
	\label{tab:tag_ranks}
	\centering
	\begin{tabular}{ r l r } 
		\toprule
		Rank & Tag & \# of Posts \\ 
		\midrule
		1 & javascript  & 2,585,182 \\
		8 & html & 1,279,137 \\
		73 & apache & 99,377 \\
		751 & web-config & 10,056 \\
		4,508 & simplify & 1,000 \\
		16,986 & against & 100 \\
		46,663 & db-charmer & 1\\
		\bottomrule
	\end{tabular}
\end{table}

After filtering the data for the snippet length and score thresholds, one problem remained with the set of valid labels. Because users are allowed to enter any text as a tag for their posts, there is a long-tailed distribution of tags that are rarely used. Table \ref{tab:tag_ranks} displays the magnitude of the problem. In the first 4,508 tags, the amount of posts per tag drops from 2.5 million to just 1,000. In order to enable a 99\% / 1\% training/test split and still have 10 positive labels per tag to estimate performance, we cut off tags with fewer than one thousand positive samples.

In the following section, we explain how we construct our models and perform validation using the snippet, score, and tag-filtered data.

\section{Methodology} \label{methodology}

Our motivations for using neural networks in this work are severalfold. As discussed in the introduction, convolutional neural networks have shown state-of-the-art performance in natural language tasks with less computation than LSTMs \cite{character_aware} \cite{nn_models}. Both natural language and source code tasks must model structure, semantic meaning, and context. 

Neural networks also have the ability to efficiently handle multilabel classification problems: rather than training \(M\) classifiers for \(M\) different output labels, the output layer of a neural network can have \(M\) nodes, simultaneously providing predictions for multiple labels. This enables the neural network to learn features that are common across labels, whereas individual classifiers must learn those relationships separately. 

\subsection{Neural Network Architecture}

\begin{figure}[t]
	\centering
	\includegraphics[width=.9\linewidth]{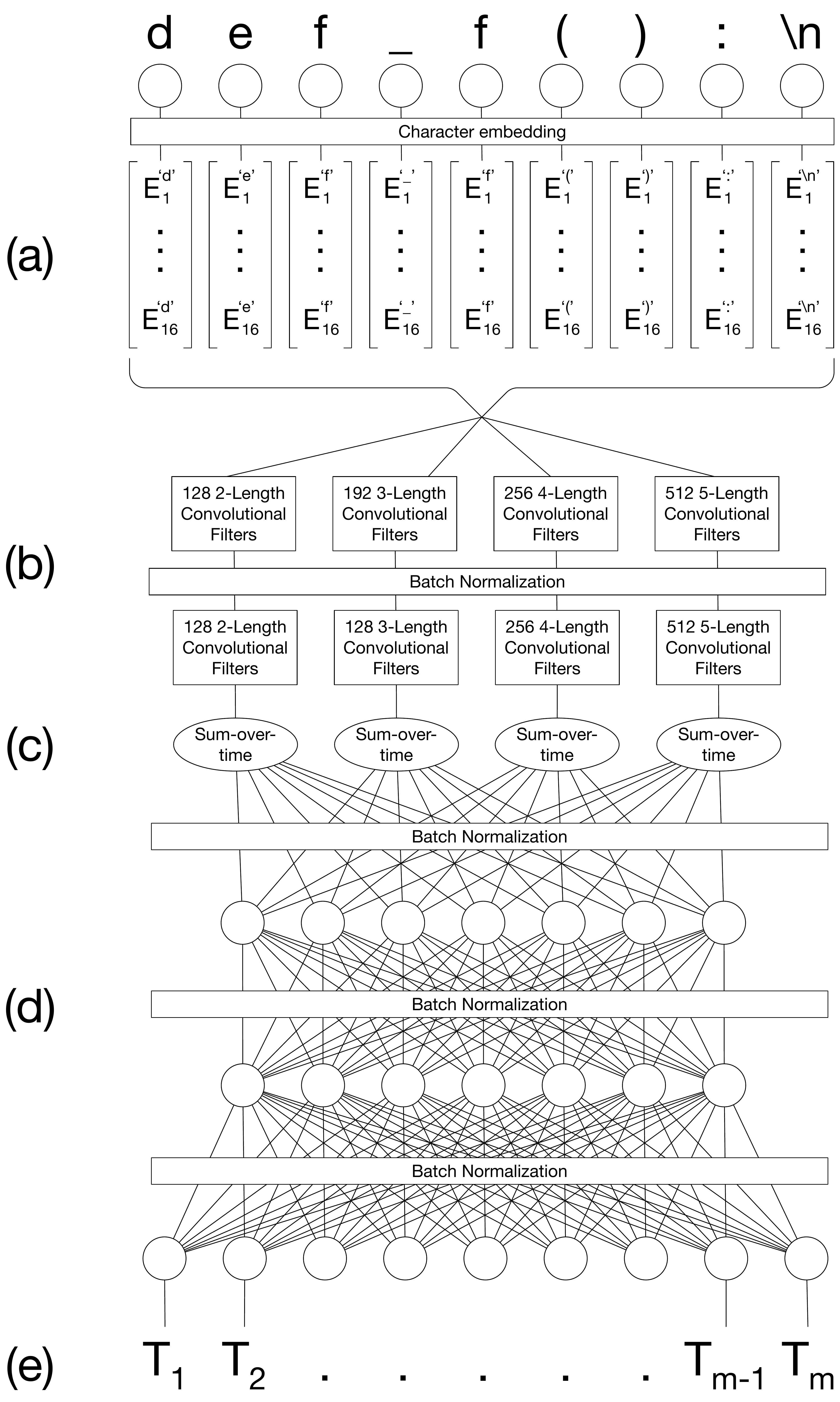}
	\caption{An overview of the neural network architecture. (a) The characters from a given code snippet are converted to real-valued vectors using a \emph{character embedding}. (b) We use stacked convolutional filters of different lengths with ReLU activations over the matrix of embeddings. (c) We perform sum-over-time pooling on the output of each stacked convolution. (d) A flattened vector is fed into two fully-connected, batch-normalized, dense layers with ReLU activations. (e) Each output node uses a logistic activation function to produce a value from 0 to 1, representing the probability of a given label.}
	\label{fig:arch}
\end{figure}

Fig. \ref{fig:arch} gives an overview of the neural network architecture. In part (a) of Fig. \ref{fig:arch}, we use a character embedding to transform each printable character into a 16-dimensional real-valued vector. We chose character embeddings over more commonly used word embeddings for multiple reasons. Creating an embedding for every word in the source code domain is problematic because of the massive set of unique identifiers. Forming a dictionary from words only seen in the training set will not generalize, and using all possible identifiers will be infeasible to optimize. The neural network only needs to optimize 100 embeddings when using the printable characters. Additionally, the character embeddings are able to function on any text, allowing the model to predict on source code without the use of language-specific parsers or features.

\begin{figure}[t]
	\centering
	\includegraphics[width=\linewidth]{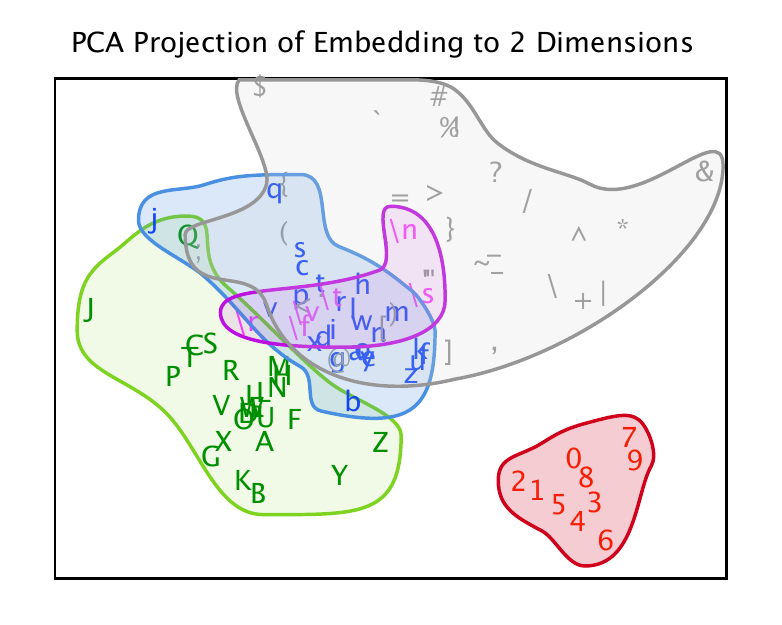}
	\caption{A two-dimensional projection of the character embedding vectors that are optimized during model training. The model generates clear spatial relationships between various sets of characters. The separation between uppercase letters, lowercase letters, symbols, and numbers is of particular interest. In general, meaningful spatial relationships significantly improve the features extracted by the convolutional layers.}
	\label{fig:embedding}
\end{figure}

In order to provide an intuition of the character embedding, we use PCA to project the 16-dimensional embedding vectors down to two dimensions, as displayed in Fig. \ref{fig:embedding}. This figure indicates that the model generates salient spatial relationships between the embedded characters during optimization, which is critical to the performance of the convolutional layers. The convolutions are able to preserve information about words and sequences by sliding over the embedding vectors of consecutive characters. We stack two convolutional layers of the same size for various filter lengths, which generates a stacked convolution matrix. Using sum-over-time pooling on the stacked convolution matrix allows us to obtain a fixed-length vector regardless of the initial input size. 

After two batch normalized dense layers, the last layer has a logistic activation for each neuron in order to output the probability of a tag occurring. The network is trained on binary vectors containing a 1 for every tag that occurs for a given code snippet and 0 otherwise.  We use binary cross-entropy as our loss function.

\subsection{Validation Setup}

Since we train the model on Stack Overflow and predict on arbitrary source code, we must validate the model in both domains. On the SO data, we use a hold-out test-set strategy so that the model can be evaluated on previously unseen data. In the source code domain, we perform human validation to verify the accuracy of the model’s outputs.

\subsection{Stack Overflow Validation}

To validate the neural network on Stack Overflow, we tested a number of multilabel test set stratification algorithms. Stratification based on k-fold cross-validation, which is a standard technique for binary and multiclass classification tasks, cannot be directly applied to the SO multilabel classification problem due to classes not being disjoint. Furthermore, due to the class imbalance caused by using a long-tailed tag distribution for labels, random stratification produces partitions of the data that do not generate good estimates for multilabel problems \cite{stratification} \cite{multi_mining}. In particular, the label counts for the top tag and the 4,508th tag differ by 3 orders of magnitude, which can result in classes with very few positive labels for the test set. 

Since deep CNN models take a long time to train and benefit from large datasets, we want to avoid cross validation and use as much of the dataset as possible to train our model. Our goal is to generate a 98\% / 1\% / 1\% train/validation/test split that still provides a good estimate of performance. With an ideal stratification, this would ensure that even the rarest tags (with 1000 samples each) would have 10 samples in the validation and test sets, which is sufficient for estimating performance. On our dataset, this would result in about 240,000 samples in validation and test sets.

Multilabel stratification begins with the $m$-by-$q$ label matrix $Y$, where $m$ is the number of samples in the dataset $D$, $q$ is the number of labels in the set of labels $\Lambda$, and $Y[i,j] = 1$ where sample $i$ has label $j$, and $Y[i,j] = 0$ otherwise.  The goal is to generate a partition $\{D^1, \ldots, D^k\}$ of $D$ that fulfills certain desirable properties. First, the size of each partition should match a designated ratio of samples, in our case, $\frac{|D^{\text{train}}|, |D^{\text{test}}|, |D^{\text{val}}|}{|D|} = (0.98, 0.1, 0.1)$. Additionally, the proportion of positive examples of each label in each partition should be the same; i.e.,

$$\forall s \in \{train, test, val\}, \forall j \in \Lambda: \frac{\sum_{i \in D^s} Y[i, j]}{|D^s|} = c_j$$ where $c_j$ is the proportion of positive examples of label $j$ in $D$. 

Labelset stratification \cite{multi_mining} considers each combination of labels, denoted labelsets, as a unique label and then performs standard k-fold stratification on those labelsets. This works well for multilabel problems where each labelset appears sufficiently often. However, this does not optimize for individual label counts, which is a problem for datasets like SO that include rare labels and rare label combinations. We found that iterative stratification \cite{stratification}, a greedy method that specifically focuses on balancing the label counts for rare labels, produced the best validation and test sets. To produce our partition, we ran iterative stratification twice with a 99\%/1\% split, which resulted in a 98.01\%/0.99\%/ 0.1\% train/validation/test split.

\begin{figure}[t]
	\centering
	\includegraphics[width=.95\linewidth]{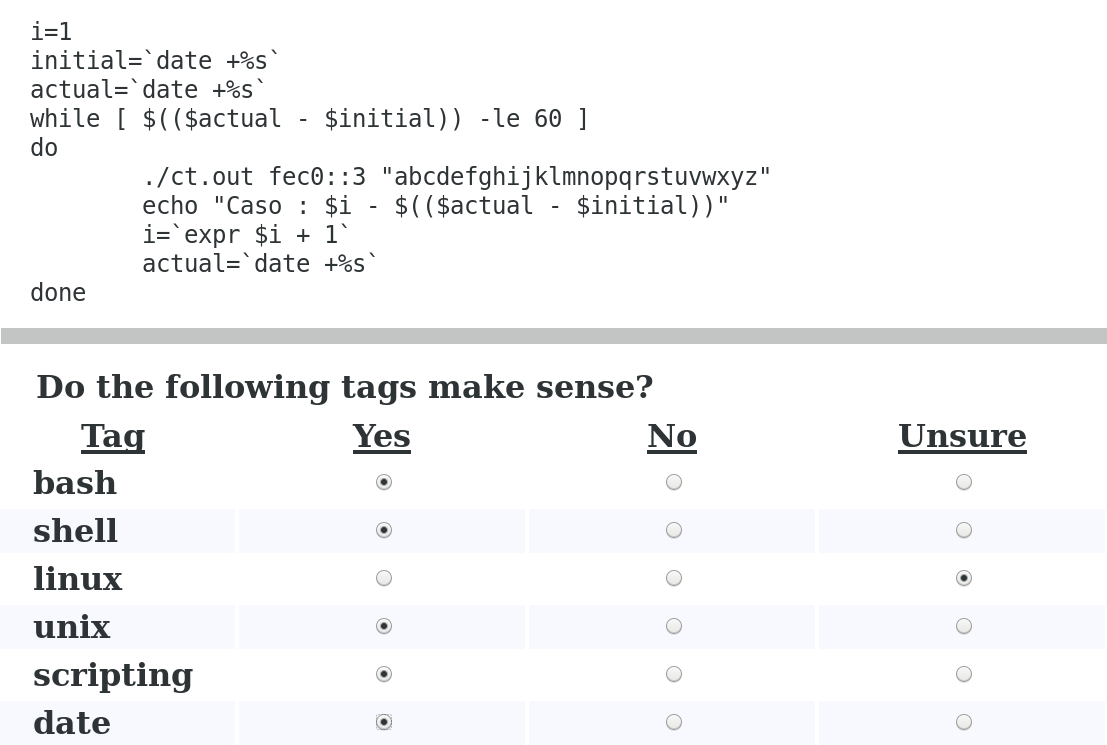}
	\caption{The GUI for human validation of model outputs on source code documents.}
	\label{fig:source_ui}
\end{figure}

\subsection{Source Code Validation}

Validating the model's performance on source code poses a different challenge because of the lack of labeled documents. In order to obtain results, we performed human validation on source code that is randomly sampled from GitHub \cite{github}. Specifically, we ran a script to download the master branches of random GitHub projects via the site's API until we had 146,408 individual files. We sampled 20 files for each of the following extensions, resulting in a total of 200 source code documents: [py, xml, java, html, c, js, sql, asm, sh, cpp]. Note that the extensions were not presented to the users and that they do not inform the predictions of the model. We created a GUI, displayed in Fig. \ref{fig:source_ui}, that presents the top labels and asks users if they agree with, disagree with, or are unsure about each label. There were a total of 3 reviewers, each of whom answered the questions on the GUI for all 200 source code documents. We remove the unsure answers and use simple voting among the remaining ratings to produce ground truth and compute an ROC curve. 

\section{Results} \label{results}

On the Stack Overflow data, we first calculated the top-1 accuracy previously used by Kuo \cite{coocc_pred} and Clayton and Byrne \cite{predicting_tags}. We obtain a 78.7\% top-1 accuracy, which is a significant improvement over the previous best of 65\%.

\begin{figure}[t]
	\centering
	\includegraphics[width=\linewidth]{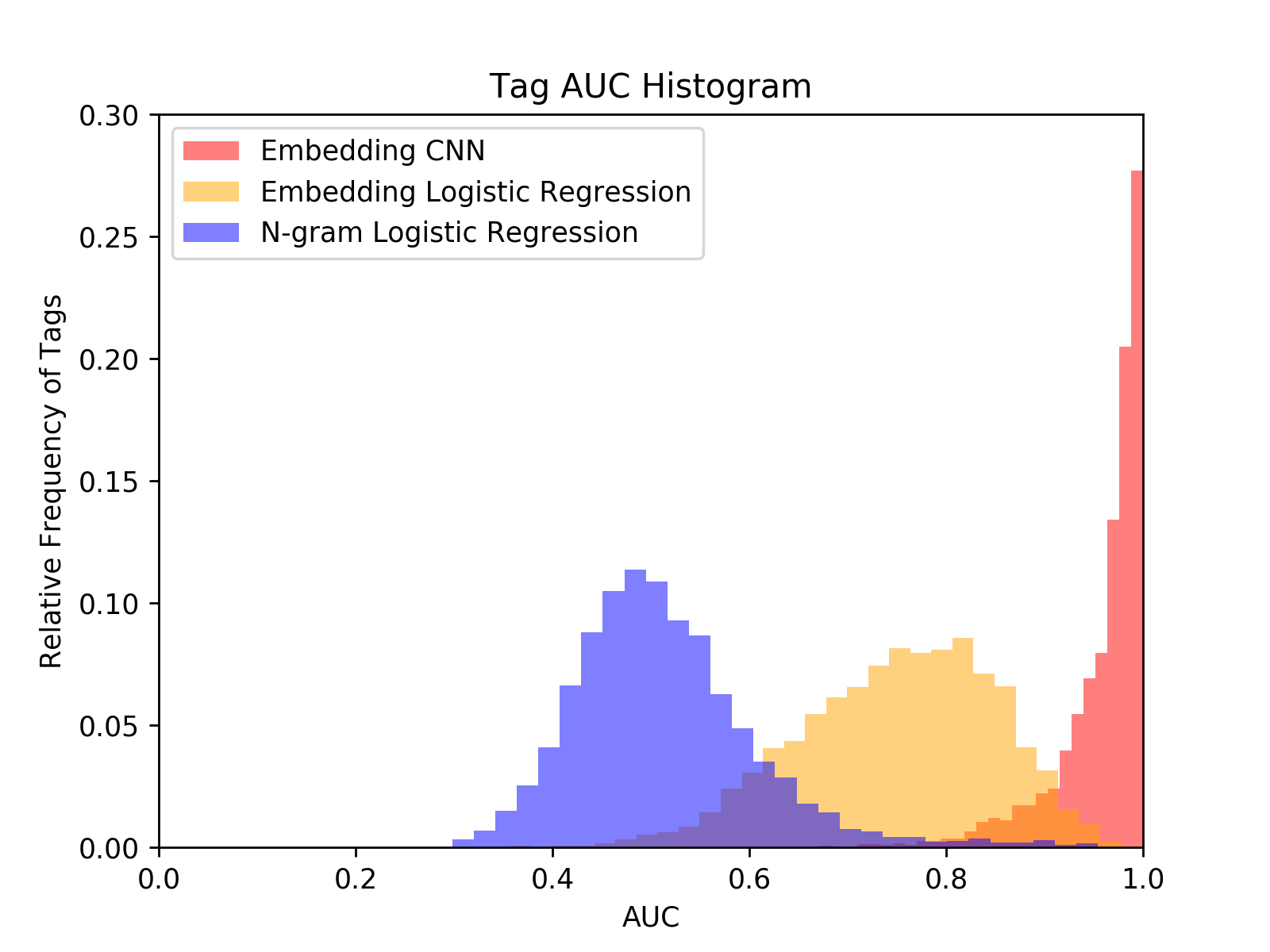}
	\caption{The distribution of tag AUCs for each model. Because our dataset uses 4,508 labels, there are 4,508 AUCs binned and plotted for each model. This graph demonstrates how well each model performs across all the labels.}
	\label{fig:auc_hist}
\end{figure}

However, we found that metric to be lacking: it only checks if the model's top prediction is in the SO post's tag set. Our goal is to predict many tags pertinent to a source code document, not just its primary tag.  Because our work is introducing the multilabel tag prediction problem on Stack Overflow code snippets, we train multiple baseline models to demonstrate the significance of our convolutional neural network architecture. In order to evaluate the results, we computed the area under ROC (AUC) for each individual tag. This is a reasonable evaluation because it demonstrates the performance of each model across the entire set of tags. 

\begin{table}[t]
	\caption{Mean, median, and standard deviation of tag AUCs for each model. }
	\label{tab:auc_statistics}
	\centering
	\begin{tabular}{ l c c c } 
		\toprule
		Model & Mean & Median & Stdev \\ 
		\midrule
		Embedding CNN & 0.957  & 0.974 & 0.048 \\
		Embedding Logistic Regression & 0.751  & 0.759 &  0.099 \\
		N-gram Logistic Regression & 0.514  & 0.502 &  0.093 \\
		\bottomrule
	\end{tabular}
\end{table}

We used two additional models as baselines for this problem. The first model performs logistic regression on a bag of n-grams. This model obtains the 75,000 most common n-grams (using n=1,2,3) from the training set to use as features. The second model performs logistic regression on a character embedding of the input code using an embedding dimension of 8. We choose these two models as baselines because they test two different types of featurizations and they are able to efficiently train and predict on multilabel problems. 

Fig. \ref{fig:auc_hist} shows the distributions of tag AUCs for the CNN model and the logistic regression baseline models. Because our dataset uses 4,508 tags, there are 4,508 AUC values that are binned and plotted for each model. The shape of the logistic regression distributions are similar - most of the tags fall within the central range of the models' distributions and there are few tags that perform relatively well or relatively poorly. Our convolutional architecture performs well on most of the tags, and instead has a long-tailed distribution of decreasing performance. 

Table \ref{tab:auc_statistics} displays a summarized, quantitative view of the tag AUC distributions. The logistic regression models have similar standard deviations, but the n-gram model has a considerably lower mean and median, indicating that the n-gram features are not as effective as the character embeddings. The convolutional network has a significantly higher mean and median, and a lower standard deviation. Although all of the models perform worse as the rarity of the tags increases, the lower standard deviation of the convolutional network implies that the model is more robust to the rarity of a given tag. 

\begin{figure}[t]
	\centering
	\includegraphics[width=\linewidth]{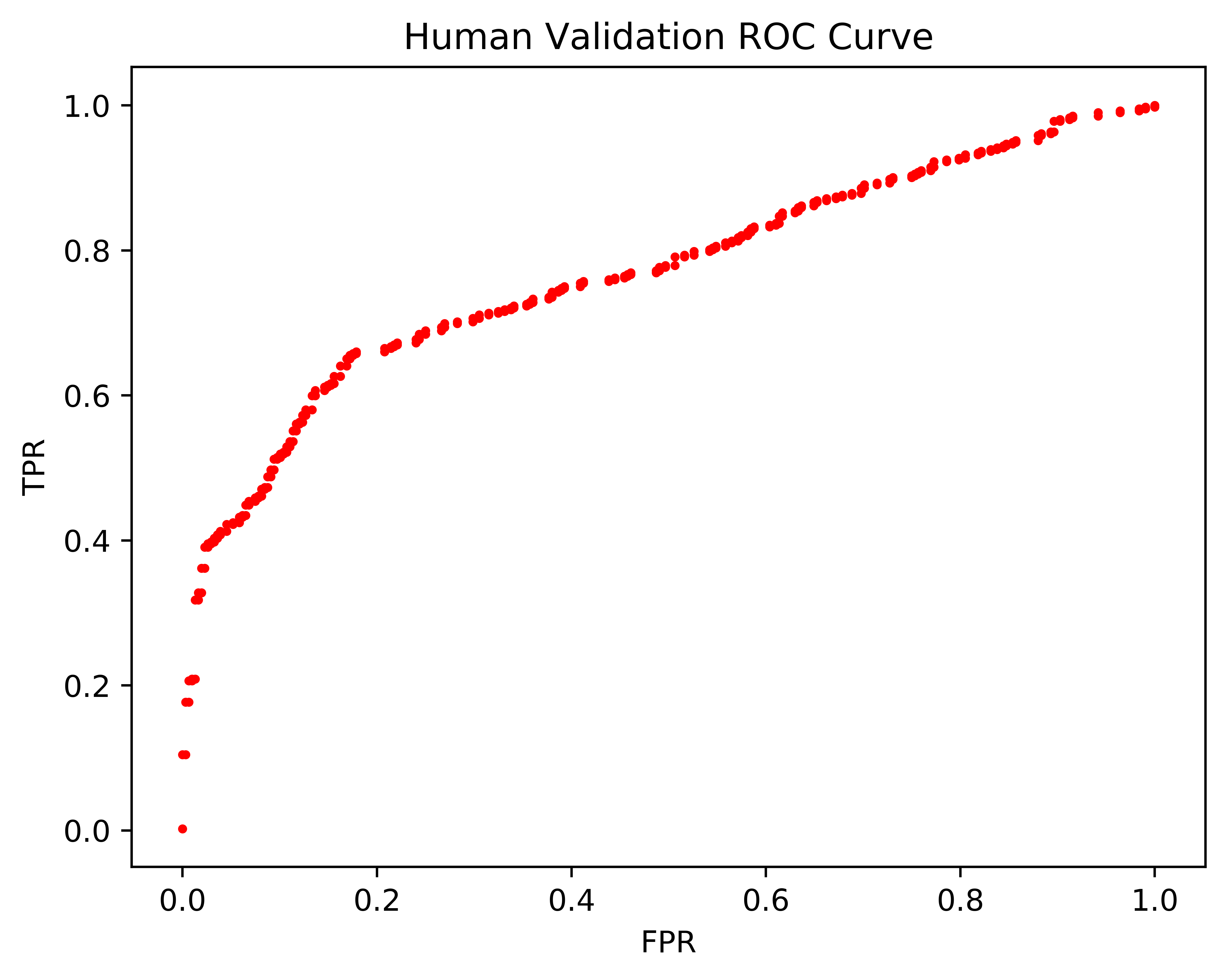}
	\caption{Human validation ROC curve with a 0.769 AUC. This differs from the Stack Overflow AUC values because it operates on the results of human validation, which is limited to only a few tags per document.}
	\label{fig:roc_m1}
\end{figure}

For source code validation, we use human feedback on the convolutional network to generate Fig. \ref{fig:roc_m1}. The model obtained a 0.769 AUC. For the sake of comparison, we compute top-1 accuracy with the human validation on source code and obtain an 86.6\% accuracy. We note that this is better than the analogous performance on Stack Overflow, which indicates that, on source code, the model performs better for the first tag, but worse for the rest. 

As a final note on performance, we trained and tested our model using an NVIDIA 1080 GPU. Our model obtains speeds of about 317,000 characters per second. Assuming an average of 38 characters per line of code (calculated based on a random sample of source files from GitHub\cite{github}), the model is able to achieve prediction speeds of 8,342 source lines of code per second. To put this in context, it would take the model less than an hour to predict on the 20+ million lines of code in the Linux kernel. It is also readily parallelized to quickly predict across much larger source code corpora.

\section{Challenges/Limitations} \label{challenges}

In the course of our research, we encountered a few limitations that require further study. First is the transfer learning problem between Stack Overflow code snippets and source code. The lack of labeled source code prevents us from training directly on the desired domain. 

The size of SO code snippets and the maximum number of tags per post are detrimental to the model's ability to predict on arbitrarily long source code. Due to the five tags per post limit, predicting more tags will increase the model loss, resulting in predictions with few tags. The original hypothesis was that the model would associate few predictions with short snippets and many tags for longer snippets, but the source code evaluation did not strongly support this. Exploring approaches that utilize loss functions other than binary cross-entropy may address these tag limit problems. 

Another issue is that Stack Overflow users do not tag their code snippets directly, but rather their questions. For example, a user could post a code snippet of an XML document, ask how to parse it in Java, and tag the thread with ``XML," ``Java," and ``parse." These tags are all extremely relevant to the user's question, but they do not describe the code snippet independently. During training, our model is only able to see that the XML document is an example of XML, Java, and parsing. This creates noise in the Java and parse labels.

Finally, the human verification process is a noisy evaluation of the model's performance on source code. Verifying the predictions is an arduous process because the model is familiar with thousands of functionalities. It is infeasible for individuals to be masters of such a wide range of ideas and tools, which results in a significant amount of labeler disagreement. 

\section{Conclusions/Future Work} \label{conclusions}

We leverage the crowdsourced data from Stack Overflow to train a deep convolutional neural network that can attach meaningful, semantic labels to source code documents of arbitrary language. While most current code search approaches locate documents by matching strings from user queries, our approach enables us to identify documents based on functional content instead of the literal characters used in source code or documentation. A logical next step is to apply this model to large source code corpora and build a search interface to find source code of interest.

Unlike previous supervised SO tag-prediction models, we train and test strictly on code snippets, yet we still advance the top-1 prediction accuracy from 65\% to 79\% on Stack Overflow. We also achieve 87\% on human-validated source code.  Using the area under ROC to measure performance, we obtain a mean AUC of 0.957 on the Stack Overflow dataset and an AUC of 0.769 on the human source code validation. Refining the methodology and data preprocessing by training the model with entire threads instead of posts could alleviate the performance drop caused by transfer learning. An alternative direction for future research is to investigate better metrics and loss functions for training and evaluating model performance on long-tailed multilabel datasets. This could prevent the model from being punished for predicting more than five tags.

Finally, extensions of the architecture that broaden the contextual aperture of the convolutional layers may grant the model a deeper understanding of abstract code concepts and semantics. This would enable more sophisticated code search and comprehension. 

\section*{Acknowledgments} \label{acknowledgements}

This project was sponsored by the Air Force Research Laboratory (AFRL) as part of the DARPA MUSE program.

	\bibliographystyle{ACM-Reference-Format}
	\bibliography{sample-bibliography} 


\begin{thebibliography}{37}

%%% the Web address.

\ifx \showCODEN    \undefined \def \showCODEN     #1{\unskip}     \fi
\ifx \showDOI      \undefined \def \showDOI       #1{#1}\fi
\ifx \showISBNx    \undefined \def \showISBNx     #1{\unskip}     \fi
\ifx \showISBNxiii \undefined \def \showISBNxiii  #1{\unskip}     \fi
\ifx \showISSN     \undefined \def \showISSN      #1{\unskip}     \fi
\ifx \showLCCN     \undefined \def \showLCCN      #1{\unskip}     \fi
\ifx \shownote     \undefined \def \shownote      #1{#1}          \fi
\ifx \showarticletitle \undefined \def \showarticletitle #1{#1}   \fi
\ifx \showURL      \undefined \def \showURL       {\relax}        \fi
% invisible to TeX
\providecommand\bibfield[2]{#2}
\providecommand\bibinfo[2]{#2}
\providecommand\natexlab[1]{#1}
\providecommand\showeprint[2][]{arXiv:#2}

\bibitem[\protect\citeauthoryear{Allamanis, Barr, Devanbu, and
  Sutton}{Allamanis et~al\mbox{.}}{2017}]%
        {surveybigcodenatural}
\bibfield{author}{\bibinfo{person}{Miltiadis Allamanis},
  \bibinfo{person}{Earl~T Barr}, \bibinfo{person}{Premkumar Devanbu}, {and}
  \bibinfo{person}{Charles Sutton}.} \bibinfo{year}{2017}\natexlab{}.
\newblock \showarticletitle{A survey of machine learning for big code and
  naturalness}.
\newblock \bibinfo{journal}{\emph{arXiv preprint arXiv:1709.06182}}
  (\bibinfo{year}{2017}).
\newblock


\bibitem[\protect\citeauthoryear{Anderson, Bothell, Byrne, Douglass, Lebiere,
  and Qin}{Anderson et~al\mbox{.}}{2004}]%
        {act-r}
\bibfield{author}{\bibinfo{person}{John~R Anderson}, \bibinfo{person}{Daniel
  Bothell}, \bibinfo{person}{Michael~D Byrne}, \bibinfo{person}{Scott
  Douglass}, \bibinfo{person}{Christian Lebiere}, {and} \bibinfo{person}{Yulin
  Qin}.} \bibinfo{year}{2004}\natexlab{}.
\newblock \showarticletitle{An integrated theory of the mind.}
\newblock \bibinfo{journal}{\emph{Psychological review}} \bibinfo{volume}{111},
  \bibinfo{number}{4} (\bibinfo{year}{2004}), \bibinfo{pages}{1036}.
\newblock


\bibitem[\protect\citeauthoryear{Bajracharya, Ngo, Linstead, Dou, Rigor, Baldi,
  and Lopes}{Bajracharya et~al\mbox{.}}{2006}]%
        {sourcerer}
\bibfield{author}{\bibinfo{person}{Sushil Bajracharya}, \bibinfo{person}{Trung
  Ngo}, \bibinfo{person}{Erik Linstead}, \bibinfo{person}{Yimeng Dou},
  \bibinfo{person}{Paul Rigor}, \bibinfo{person}{Pierre Baldi}, {and}
  \bibinfo{person}{Cristina Lopes}.} \bibinfo{year}{2006}\natexlab{}.
\newblock \showarticletitle{Sourcerer: a search engine for open source code
  supporting structure-based search}. In \bibinfo{booktitle}{\emph{Companion to
  the 21st ACM SIGPLAN symposium on Object-oriented programming systems,
  languages, and applications}}. ACM, \bibinfo{pages}{681--682}.
\newblock


\bibitem[\protect\citeauthoryear{Berry, Dumais, and O'Brien}{Berry
  et~al\mbox{.}}{1995}]%
        {lsi_search}
\bibfield{author}{\bibinfo{person}{Michael~W Berry}, \bibinfo{person}{Susan~T
  Dumais}, {and} \bibinfo{person}{Gavin~W O'Brien}.}
  \bibinfo{year}{1995}\natexlab{}.
\newblock \showarticletitle{Using linear algebra for intelligent information
  retrieval}.
\newblock \bibinfo{journal}{\emph{SIAM review}} \bibinfo{volume}{37},
  \bibinfo{number}{4} (\bibinfo{year}{1995}), \bibinfo{pages}{573--595}.
\newblock


\bibitem[\protect\citeauthoryear{{Black Duck}}{{Black Duck}}{2017}]%
        {openhub}
\bibfield{author}{\bibinfo{person}{{Black Duck}}.}
  \bibinfo{year}{2017}\natexlab{}.
\newblock \bibinfo{title}{Open Hub}.
\newblock
\newblock
\urldef\tempurl%
\url{https://www.openhub.net}
\showURL{%
\tempurl}


\bibitem[\protect\citeauthoryear{Charte, Rivera, del Jesus, and Herrera}{Charte
  et~al\mbox{.}}{2015}]%
        {irbl}
\bibfield{author}{\bibinfo{person}{Francisco Charte},
  \bibinfo{person}{Antonio~J Rivera}, \bibinfo{person}{Mar{\'\i}a~J del Jesus},
  {and} \bibinfo{person}{Francisco Herrera}.} \bibinfo{year}{2015}\natexlab{}.
\newblock \showarticletitle{Addressing imbalance in multilabel classification:
  Measures and random resampling algorithms}.
\newblock \bibinfo{journal}{\emph{Neurocomputing}}  \bibinfo{volume}{163}
  (\bibinfo{year}{2015}), \bibinfo{pages}{3--16}.
\newblock


\bibitem[\protect\citeauthoryear{Dam, Tran, and Pham}{Dam
  et~al\mbox{.}}{2016}]%
        {dam2016deep}
\bibfield{author}{\bibinfo{person}{Hoa~Khanh Dam}, \bibinfo{person}{Truyen
  Tran}, {and} \bibinfo{person}{Trang Pham}.} \bibinfo{year}{2016}\natexlab{}.
\newblock \showarticletitle{A deep language model for software code}.
\newblock \bibinfo{journal}{\emph{arXiv preprint arXiv:1608.02715}}
  (\bibinfo{year}{2016}).
\newblock


\bibitem[\protect\citeauthoryear{Dauphin, Fan, Auli, and Grangier}{Dauphin
  et~al\mbox{.}}{2016}]%
        {nn_models}
\bibfield{author}{\bibinfo{person}{Yann~N Dauphin}, \bibinfo{person}{Angela
  Fan}, \bibinfo{person}{Michael Auli}, {and} \bibinfo{person}{David
  Grangier}.} \bibinfo{year}{2016}\natexlab{}.
\newblock \showarticletitle{Language Modeling with Gated Convolutional
  Networks}.
\newblock \bibinfo{journal}{\emph{arXiv preprint arXiv:1612.08083}}
  (\bibinfo{year}{2016}).
\newblock


\bibitem[\protect\citeauthoryear{De~Lucia, Di~Penta, Oliveto, Panichella, and
  Panichella}{De~Lucia et~al\mbox{.}}{2014}]%
        {labeling_review}
\bibfield{author}{\bibinfo{person}{Andrea De~Lucia},
  \bibinfo{person}{Massimiliano Di~Penta}, \bibinfo{person}{Rocco Oliveto},
  \bibinfo{person}{Annibale Panichella}, {and} \bibinfo{person}{Sebastiano
  Panichella}.} \bibinfo{year}{2014}\natexlab{}.
\newblock \showarticletitle{Labeling source code with information retrieval
  methods: an empirical study}.
\newblock \bibinfo{journal}{\emph{Empirical Software Engineering}}
  \bibinfo{volume}{19}, \bibinfo{number}{5} (\bibinfo{year}{2014}),
  \bibinfo{pages}{1383--1420}.
\newblock


\bibitem[\protect\citeauthoryear{Deshpande and Riehle}{Deshpande and
  Riehle}{2008}]%
        {source_growth}
\bibfield{author}{\bibinfo{person}{Amit Deshpande} {and} \bibinfo{person}{Dirk
  Riehle}.} \bibinfo{year}{2008}\natexlab{}.
\newblock \showarticletitle{The total growth of open source}. In
  \bibinfo{booktitle}{\emph{IFIP International Conference on Open Source
  Systems}}. Springer, \bibinfo{pages}{197--209}.
\newblock


\bibitem[\protect\citeauthoryear{Gethers, Savage, Di~Penta, Oliveto,
  Poshyvanyk, and De~Lucia}{Gethers et~al\mbox{.}}{2011}]%
        {relational_topics}
\bibfield{author}{\bibinfo{person}{Malcom Gethers}, \bibinfo{person}{Trevor
  Savage}, \bibinfo{person}{Massimiliano Di~Penta}, \bibinfo{person}{Rocco
  Oliveto}, \bibinfo{person}{Denys Poshyvanyk}, {and} \bibinfo{person}{Andrea
  De~Lucia}.} \bibinfo{year}{2011}\natexlab{}.
\newblock \showarticletitle{CodeTopics: which topic am I coding now?}. In
  \bibinfo{booktitle}{\emph{Proceedings of the 33rd International Conference on
  Software Engineering}}. ACM, \bibinfo{pages}{1034--1036}.
\newblock


\bibitem[\protect\citeauthoryear{{GitHub}}{{GitHub}}{2017}]%
        {github}
\bibfield{author}{\bibinfo{person}{{GitHub}}.} \bibinfo{year}{2017}\natexlab{}.
\newblock \bibinfo{title}{GitHub}.
\newblock
\newblock
\urldef\tempurl%
\url{https://github.com}
\showURL{%
\tempurl}


\bibitem[\protect\citeauthoryear{Grabowski, Kruszewska, and
  Kosi{\'n}ski}{Grabowski et~al\mbox{.}}{2008}]%
        {onlineproperties}
\bibfield{author}{\bibinfo{person}{A Grabowski}, \bibinfo{person}{N
  Kruszewska}, {and} \bibinfo{person}{RA Kosi{\'n}ski}.}
  \bibinfo{year}{2008}\natexlab{}.
\newblock \showarticletitle{Properties of on-line social systems}.
\newblock \bibinfo{journal}{\emph{The European Physical Journal B-Condensed
  Matter and Complex Systems}} \bibinfo{volume}{66}, \bibinfo{number}{1}
  (\bibinfo{year}{2008}), \bibinfo{pages}{107--113}.
\newblock


\bibitem[\protect\citeauthoryear{Hindle, Barr, Su, Gabel, and Devanbu}{Hindle
  et~al\mbox{.}}{2012}]%
        {naturalness}
\bibfield{author}{\bibinfo{person}{Abram Hindle}, \bibinfo{person}{Earl~T
  Barr}, \bibinfo{person}{Zhendong Su}, \bibinfo{person}{Mark Gabel}, {and}
  \bibinfo{person}{Premkumar Devanbu}.} \bibinfo{year}{2012}\natexlab{}.
\newblock \showarticletitle{On the naturalness of software}. In
  \bibinfo{booktitle}{\emph{Software Engineering (ICSE), 2012 34th
  International Conference on}}. IEEE, \bibinfo{pages}{837--847}.
\newblock


\bibitem[\protect\citeauthoryear{Hochreiter and Schmidhuber}{Hochreiter and
  Schmidhuber}{1997}]%
        {original_lstm}
\bibfield{author}{\bibinfo{person}{Sepp Hochreiter} {and}
  \bibinfo{person}{J{\"u}rgen Schmidhuber}.} \bibinfo{year}{1997}\natexlab{}.
\newblock \showarticletitle{Long short-term memory}.
\newblock \bibinfo{journal}{\emph{Neural computation}} \bibinfo{volume}{9},
  \bibinfo{number}{8} (\bibinfo{year}{1997}), \bibinfo{pages}{1735--1780}.
\newblock


\bibitem[\protect\citeauthoryear{Kim, Jernite, Sontag, and Rush}{Kim
  et~al\mbox{.}}{2015}]%
        {character_aware}
\bibfield{author}{\bibinfo{person}{Yoon Kim}, \bibinfo{person}{Yacine Jernite},
  \bibinfo{person}{David Sontag}, {and} \bibinfo{person}{Alexander~M Rush}.}
  \bibinfo{year}{2015}\natexlab{}.
\newblock \showarticletitle{Character-aware neural language models}.
\newblock \bibinfo{journal}{\emph{arXiv preprint arXiv:1508.06615}}
  (\bibinfo{year}{2015}).
\newblock


\bibitem[\protect\citeauthoryear{{Krugle}}{{Krugle}}{2017}]%
        {krugle}
\bibfield{author}{\bibinfo{person}{{Krugle}}.} \bibinfo{year}{2017}\natexlab{}.
\newblock \bibinfo{title}{krugle}.
\newblock
\newblock
\urldef\tempurl%
\url{http://opensearch.krugle.org}
\showURL{%
\tempurl}


\bibitem[\protect\citeauthoryear{Kuhn, Ducasse, and G{\'\i}rba}{Kuhn
  et~al\mbox{.}}{2007}]%
        {semantic_clustering}
\bibfield{author}{\bibinfo{person}{Adrian Kuhn}, \bibinfo{person}{St{\'e}phane
  Ducasse}, {and} \bibinfo{person}{Tudor G{\'\i}rba}.}
  \bibinfo{year}{2007}\natexlab{}.
\newblock \showarticletitle{Semantic clustering: Identifying topics in source
  code}.
\newblock \bibinfo{journal}{\emph{Information and Software Technology}}
  \bibinfo{volume}{49}, \bibinfo{number}{3} (\bibinfo{year}{2007}),
  \bibinfo{pages}{230--243}.
\newblock


\bibitem[\protect\citeauthoryear{Kuo}{Kuo}{2011}]%
        {coocc_pred}
\bibfield{author}{\bibinfo{person}{Darren Kuo}.}
  \bibinfo{year}{2011}\natexlab{}.
\newblock \bibinfo{booktitle}{\emph{On word prediction methods}}.
\newblock \bibinfo{type}{{T}echnical {R}eport}. \bibinfo{institution}{Technical
  report, EECS Department, University of California, Berkeley}.
\newblock


\bibitem[\protect\citeauthoryear{Lazzarini~Lemos, Bajracharya, and
  Ossher}{Lazzarini~Lemos et~al\mbox{.}}{2007}]%
        {codegenie}
\bibfield{author}{\bibinfo{person}{Ot{\'a}vio~Augusto Lazzarini~Lemos},
  \bibinfo{person}{Sushil~Krishna Bajracharya}, {and} \bibinfo{person}{Joel
  Ossher}.} \bibinfo{year}{2007}\natexlab{}.
\newblock \showarticletitle{CodeGenie:: a tool for test-driven source code
  search}. In \bibinfo{booktitle}{\emph{Companion to the 22nd ACM SIGPLAN
  conference on Object-oriented programming systems and applications
  companion}}. ACM, \bibinfo{pages}{917--918}.
\newblock


\bibitem[\protect\citeauthoryear{Lientz, Swanson, and Tompkins}{Lientz
  et~al\mbox{.}}{1978}]%
        {cost_main_evol}
\bibfield{author}{\bibinfo{person}{Bennet~P Lientz}, \bibinfo{person}{E.~Burton
  Swanson}, {and} \bibinfo{person}{Gail~E Tompkins}.}
  \bibinfo{year}{1978}\natexlab{}.
\newblock \showarticletitle{Characteristics of application software
  maintenance}.
\newblock \bibinfo{journal}{\emph{Commun. ACM}} \bibinfo{volume}{21},
  \bibinfo{number}{6} (\bibinfo{year}{1978}), \bibinfo{pages}{466--471}.
\newblock


\bibitem[\protect\citeauthoryear{Maarek, Berry, and Kaiser}{Maarek
  et~al\mbox{.}}{1991}]%
        {ir_libraries}
\bibfield{author}{\bibinfo{person}{Yoelle~S Maarek}, \bibinfo{person}{Daniel~M
  Berry}, {and} \bibinfo{person}{Gail~E Kaiser}.}
  \bibinfo{year}{1991}\natexlab{}.
\newblock \showarticletitle{An information retrieval approach for automatically
  constructing software libraries}.
\newblock \bibinfo{journal}{\emph{IEEE Transactions on software Engineering}}
  \bibinfo{volume}{17}, \bibinfo{number}{8} (\bibinfo{year}{1991}),
  \bibinfo{pages}{800--813}.
\newblock


\bibitem[\protect\citeauthoryear{Mcauliffe and Blei}{Mcauliffe and
  Blei}{2008}]%
        {slda}
\bibfield{author}{\bibinfo{person}{Jon~D Mcauliffe} {and}
  \bibinfo{person}{David~M Blei}.} \bibinfo{year}{2008}\natexlab{}.
\newblock \showarticletitle{Supervised topic models}. In
  \bibinfo{booktitle}{\emph{Advances in neural information processing
  systems}}. \bibinfo{pages}{121--128}.
\newblock


\bibitem[\protect\citeauthoryear{McMillan, Grechanik, Poshyvanyk, Fu, and
  Xie}{McMillan et~al\mbox{.}}{2012}]%
        {exemplar}
\bibfield{author}{\bibinfo{person}{Collin McMillan}, \bibinfo{person}{Mark
  Grechanik}, \bibinfo{person}{Denys Poshyvanyk}, \bibinfo{person}{Chen Fu},
  {and} \bibinfo{person}{Qing Xie}.} \bibinfo{year}{2012}\natexlab{}.
\newblock \showarticletitle{Exemplar: A source code search engine for finding
  highly relevant applications}.
\newblock \bibinfo{journal}{\emph{IEEE Transactions on Software Engineering}}
  \bibinfo{volume}{38}, \bibinfo{number}{5} (\bibinfo{year}{2012}),
  \bibinfo{pages}{1069--1087}.
\newblock


\bibitem[\protect\citeauthoryear{Overflow}{Overflow}{2017}]%
        {stackoverflow}
\bibfield{author}{\bibinfo{person}{Stack Overflow}.}
  \bibinfo{year}{2017}\natexlab{}.
\newblock \bibinfo{title}{Stack Overflow}.
\newblock
\newblock
\urldef\tempurl%
\url{http://stackoverflow.com}
\showURL{%
\tempurl}


\bibitem[\protect\citeauthoryear{Paul and Prakash}{Paul and Prakash}{1994}]%
        {patterns}
\bibfield{author}{\bibinfo{person}{Santanu Paul} {and} \bibinfo{person}{Atul
  Prakash}.} \bibinfo{year}{1994}\natexlab{}.
\newblock \showarticletitle{A framework for source code search using program
  patterns}.
\newblock \bibinfo{journal}{\emph{IEEE Transactions on Software Engineering}}
  \bibinfo{volume}{20}, \bibinfo{number}{6} (\bibinfo{year}{1994}),
  \bibinfo{pages}{463--475}.
\newblock


\bibitem[\protect\citeauthoryear{Reiss}{Reiss}{2009}]%
        {semantics}
\bibfield{author}{\bibinfo{person}{Steven~P Reiss}.}
  \bibinfo{year}{2009}\natexlab{}.
\newblock \showarticletitle{Semantics-based code search}. In
  \bibinfo{booktitle}{\emph{Proceedings of the 31st International Conference on
  Software Engineering}}. IEEE Computer Society, \bibinfo{pages}{243--253}.
\newblock


\bibitem[\protect\citeauthoryear{Saxe, Turner, and Blokhin}{Saxe
  et~al\mbox{.}}{2014}]%
        {crowd}
\bibfield{author}{\bibinfo{person}{Joshua Saxe}, \bibinfo{person}{Rafael
  Turner}, {and} \bibinfo{person}{Kristina Blokhin}.}
  \bibinfo{year}{2014}\natexlab{}.
\newblock \showarticletitle{CrowdSource: Automated inference of high level
  malware functionality from low-level symbols using a crowd trained machine
  learning model}. In \bibinfo{booktitle}{\emph{Malicious and Unwanted
  Software: The Americas (MALWARE), 2014 9th International Conference on}}.
  IEEE, \bibinfo{pages}{68--75}.
\newblock


\bibitem[\protect\citeauthoryear{{searchcode}}{{searchcode}}{2017}]%
        {searchcode}
\bibfield{author}{\bibinfo{person}{{searchcode}}.}
  \bibinfo{year}{2017}\natexlab{}.
\newblock \bibinfo{title}{searchcode}.
\newblock
\newblock
\urldef\tempurl%
\url{https://searchcode.com}
\showURL{%
\tempurl}


\bibitem[\protect\citeauthoryear{Sechidis, Tsoumakas, and Vlahavas}{Sechidis
  et~al\mbox{.}}{2011}]%
        {stratification}
\bibfield{author}{\bibinfo{person}{Konstantinos Sechidis},
  \bibinfo{person}{Grigorios Tsoumakas}, {and} \bibinfo{person}{Ioannis
  Vlahavas}.} \bibinfo{year}{2011}\natexlab{}.
\newblock \showarticletitle{On the stratification of multi-label data}. In
  \bibinfo{booktitle}{\emph{Joint European Conference on Machine Learning and
  Knowledge Discovery in Databases}}. Springer, \bibinfo{pages}{145--158}.
\newblock


\bibitem[\protect\citeauthoryear{{SourceForge}}{{SourceForge}}{2017}]%
        {sourceforge}
\bibfield{author}{\bibinfo{person}{{SourceForge}}.}
  \bibinfo{year}{2017}\natexlab{}.
\newblock \bibinfo{title}{SourceForge}.
\newblock
\newblock
\urldef\tempurl%
\url{https://sourceforge.net}
\showURL{%
\tempurl}


\bibitem[\protect\citeauthoryear{{Sourcegraph}}{{Sourcegraph}}{2017}]%
        {sourcegraph}
\bibfield{author}{\bibinfo{person}{{Sourcegraph}}.}
  \bibinfo{year}{2017}\natexlab{}.
\newblock \bibinfo{title}{Sourcegraph}.
\newblock
\newblock
\urldef\tempurl%
\url{https://sourcegraph.com}
\showURL{%
\tempurl}


\bibitem[\protect\citeauthoryear{Stanley and Byrne}{Stanley and Byrne}{2013}]%
        {predicting_tags}
\bibfield{author}{\bibinfo{person}{Clayton Stanley} {and}
  \bibinfo{person}{Michael~D Byrne}.} \bibinfo{year}{2013}\natexlab{}.
\newblock \showarticletitle{Predicting tags for stackoverflow posts}. In
  \bibinfo{booktitle}{\emph{Proceedings of ICCM}}, Vol.~\bibinfo{volume}{2013}.
\newblock


\bibitem[\protect\citeauthoryear{Sundermeyer, Schl{\"u}ter, and
  Ney}{Sundermeyer et~al\mbox{.}}{2012}]%
        {lstm_language}
\bibfield{author}{\bibinfo{person}{Martin Sundermeyer}, \bibinfo{person}{Ralf
  Schl{\"u}ter}, {and} \bibinfo{person}{Hermann Ney}.}
  \bibinfo{year}{2012}\natexlab{}.
\newblock \showarticletitle{LSTM Neural Networks for Language Modeling.}. In
  \bibinfo{booktitle}{\emph{Interspeech}}. \bibinfo{pages}{194--197}.
\newblock


\bibitem[\protect\citeauthoryear{Thomas, Adams, Hassan, and Blostein}{Thomas
  et~al\mbox{.}}{2010}]%
        {lda_labels}
\bibfield{author}{\bibinfo{person}{Stephen~W Thomas}, \bibinfo{person}{Bram
  Adams}, \bibinfo{person}{Ahmed~E Hassan}, {and} \bibinfo{person}{Dorothea
  Blostein}.} \bibinfo{year}{2010}\natexlab{}.
\newblock \showarticletitle{Validating the use of topic models for software
  evolution}. In \bibinfo{booktitle}{\emph{Source Code Analysis and
  Manipulation (SCAM), 2010 10th IEEE Working Conference on}}. IEEE,
  \bibinfo{pages}{55--64}.
\newblock


\bibitem[\protect\citeauthoryear{Tsoumakas, Katakis, and Vlahavas}{Tsoumakas
  et~al\mbox{.}}{2009}]%
        {multi_mining}
\bibfield{author}{\bibinfo{person}{Grigorios Tsoumakas},
  \bibinfo{person}{Ioannis Katakis}, {and} \bibinfo{person}{Ioannis Vlahavas}.}
  \bibinfo{year}{2009}\natexlab{}.
\newblock \showarticletitle{Mining multi-label data}.
\newblock In \bibinfo{booktitle}{\emph{Data mining and knowledge discovery
  handbook}}. \bibinfo{publisher}{Springer}, \bibinfo{pages}{667--685}.
\newblock


\bibitem[\protect\citeauthoryear{Wu and Zhou}{Wu and Zhou}{2016}]%
        {wu2016unified}
\bibfield{author}{\bibinfo{person}{Xi-Zhu Wu} {and} \bibinfo{person}{Zhi-Hua
  Zhou}.} \bibinfo{year}{2016}\natexlab{}.
\newblock \showarticletitle{A Unified View of Multi-Label Performance
  Measures}.
\newblock \bibinfo{journal}{\emph{arXiv preprint arXiv:1609.00288}}
  (\bibinfo{year}{2016}).
\newblock


\end{thebibliography}
	\nocite{onlineproperties}
	\nocite{crowd}
	\nocite{source_growth}
	\nocite{labeling_review}
	\nocite{ir_libraries}
	\nocite{relational_topics}
	\nocite{lda_labels}
	\nocite{cost_main_evol}
	\nocite{naturalness}
	\nocite{surveybigcodenatural}
	\nocite{lsi_search}
	\nocite{coocc_pred}
	\nocite{predicting_tags}
	\nocite{act-r}
	\nocite{semantic_clustering}
	\nocite{slda}
	\nocite{stackoverflow}
	\nocite{nn_models}
	\nocite{stratification}
	\nocite{multi_mining}
	\nocite{lstm_language}
	\nocite{original_lstm}
	\nocite{irbl}
	\nocite{github}
	\nocite{codegenie}
	\nocite{sourcerer}
	\nocite{semantics}
	\nocite{patterns}
	\nocite{sourceforge}
	\nocite{exemplar}
	\nocite{wu2016unified}
	\nocite{dam2016deep}
	
\end{document}